\documentclass[10pt,twocolumn,letterpaper,pagenumbers]{article}

\usepackage{cvpr}              

\definecolor{cvprblue}{rgb}{0.21,0.49,0.74}
\usepackage[pagebackref,breaklinks,colorlinks,allcolors=cvprblue]{hyperref}

\usepackage{cuted}
\usepackage{multirow}
\usepackage[accsupp]{axessibility} 
\usepackage{bm}
\usepackage{colortbl}
\usepackage{color}
\definecolor{mygray}{gray}{.9}
\newcommand{\CC}[1]{\cellcolor{mygray}}
\def\paperID{44xx} 
\def\confName{CVPR}
\def\confYear{2025}

\begin{document}

\title{Do We Need to Design Specific Diffusion Models for Different Tasks? Try \textcolor[RGB]{191,30,46}{ONE-PIC}}

\author{Ming Tao\textsuperscript{1,2} \quad 
Bing-Kun Bao\textsuperscript{1,2}\thanks{Corresponding Author} \quad
Yaowei Wang\textsuperscript{2} \quad 
Changsheng Xu\textsuperscript{2,3,4} \\
\textsuperscript{1}Nanjing University of Posts and Telecommunications \quad  
\textsuperscript{2}Pengcheng Laboratory \quad \\
\textsuperscript{3}University of Chinese Academy of Sciences \\
\textsuperscript{4}NLPR, Institute of Automation, CAS \\
}

\maketitle

\begin{strip}
    \vspace*{-14mm}
    \centering
    \includegraphics[width=1\textwidth]{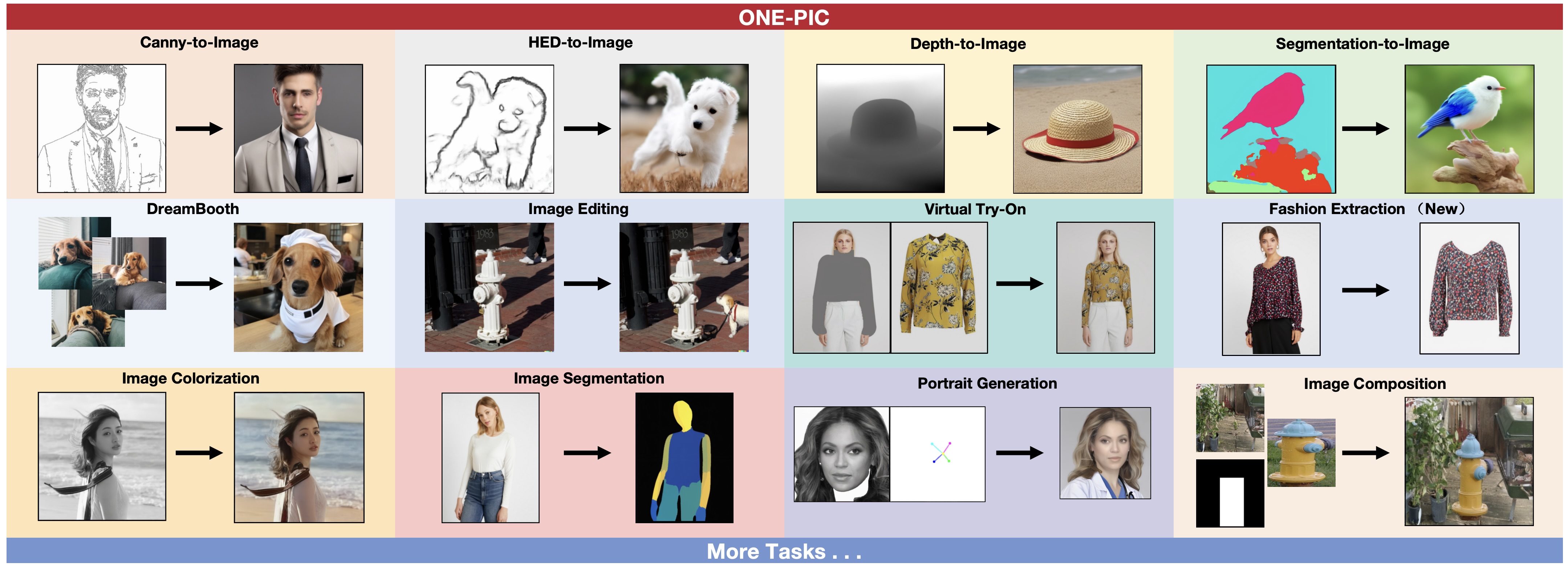}
    \captionof{figure}{Our ONE-PIC provides a simple and versatile framework for fine-tuning across various downstream image generation tasks.}
    \label{fig1}
\end{strip}

\begin{abstract}
Large pretrained diffusion models have demonstrated impressive generation capabilities and have been adapted to various downstream tasks. 
However, unlike Large Language Models (LLMs) that can learn multiple tasks in a single model based on instructed data, 
diffusion models always require additional branches, task-specific training strategies, and losses for effective adaptation to different downstream tasks. 
This task-specific fine-tuning approach brings two drawbacks.
1) The task-specific additional networks create gaps between pretraining and fine-tuning which hinders the transfer of pretrained knowledge.
2) It necessitates careful additional network design, raising the barrier to learning and implementation, and making it less user-friendly.
Thus, a question arises: Can we achieve a simple, efficient, and general approach to fine-tune diffusion models? 
To this end, we propose ONE-PIC. 
It enhances the inherited generative ability in the pretrained diffusion models without introducing additional modules. 
Specifically, we propose In-Visual-Context Tuning, 
which constructs task-specific training data by arranging source images and target images into a single image.
This approach makes downstream fine-tuning closer to the pertaining, 
allowing our model to adapt more quickly to various downstream tasks.
Moreover, we propose a Masking Strategy to unify different generative tasks. 
This strategy transforms various downstream fine-tuning tasks into predictions of the masked portions.
The extensive experimental results demonstrate that our method is simple and efficient which streamlines the adaptation process and achieves excellent performance with lower costs.
Code is available at \url{https://github.com/tobran/ONE-PIC}.
\end{abstract} 

\begin{figure*}[t]
  \centering
  \includegraphics[width=\linewidth]{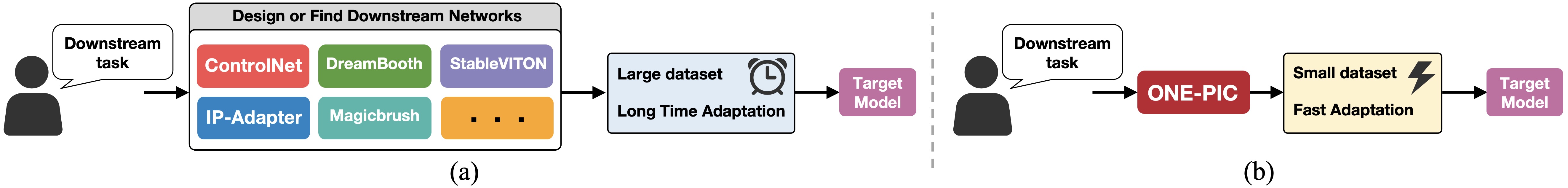}
  \caption{(a) The current adaptation process for downstream fine-tuning models in diffusion models.
(b) Our ONE-PIC significantly simplifies the network design and fine-tuning process.  
  }
  \label{fig2}
\end{figure*}

\section{Introduction}
\label{sec:intro}

In recent years, large pretrained generative models have achieved remarkable success across a variety of applications. 
One notable area in generative modeling is text-to-image synthesis, which has garnered significant attention due to its practical applications. 
This has resulted in the development of large pretrained text-to-image models, such as DALL-E~\cite{ramesh2021zero} and LDM~\cite{rombach2022high}. 
These models enhance the possibilities and capabilities of text-to-image synthesis, 
enabling the generation of visually appealing images that align well with textual descriptions.

Leveraging the powerful image-generation capabilities of pretrained diffusion models, 
recent works such as ControlNet \cite{zhang2023adding} and IP-Adapter\cite{ye2023ipadapter} have pushed the boundaries of pretrained models beyond text-to-image generation, 
and Dreambooth \cite{ruiz2023dreambooth} enables subject-driven generation.
These networks \cite{zhang2023adding,ye2023ipadapter,ruiz2023dreambooth,brooks2023instructpix2pix,kim2024stableviton} design task-specific fine-tuning frameworks based on the target of image generation. 
Although impressive results have been presented, the task-specific fine-tuning approach for large pretrained diffusion models presents several flaws. 
1) Network Design Gap: The task-specific fine-tuning models often require additional side networks to adapt to specific tasks, resulting in significant differences from the pretraining process. 
This hampers the effective utilization of pretrained knowledge and slows down the adaptation to downstream tasks.
2) Complex Model Usage and Sharing: The task-specific fine-tuning approach leads to different fine-tuned models for each specific task. 
As shown in Figure \ref{fig2}(a), models with different structures also increase the complexity of using and sharing models, raising the barrier to learning and implementation, and making it less user-friendly.
Based on the analysis provided, we can observe that the lack of a general fine-tuning approach imposes limitations on the efficiency and generalization of downstream adaptation.

\begin{figure}[t] \small
  \centering
  \includegraphics[width=\linewidth]{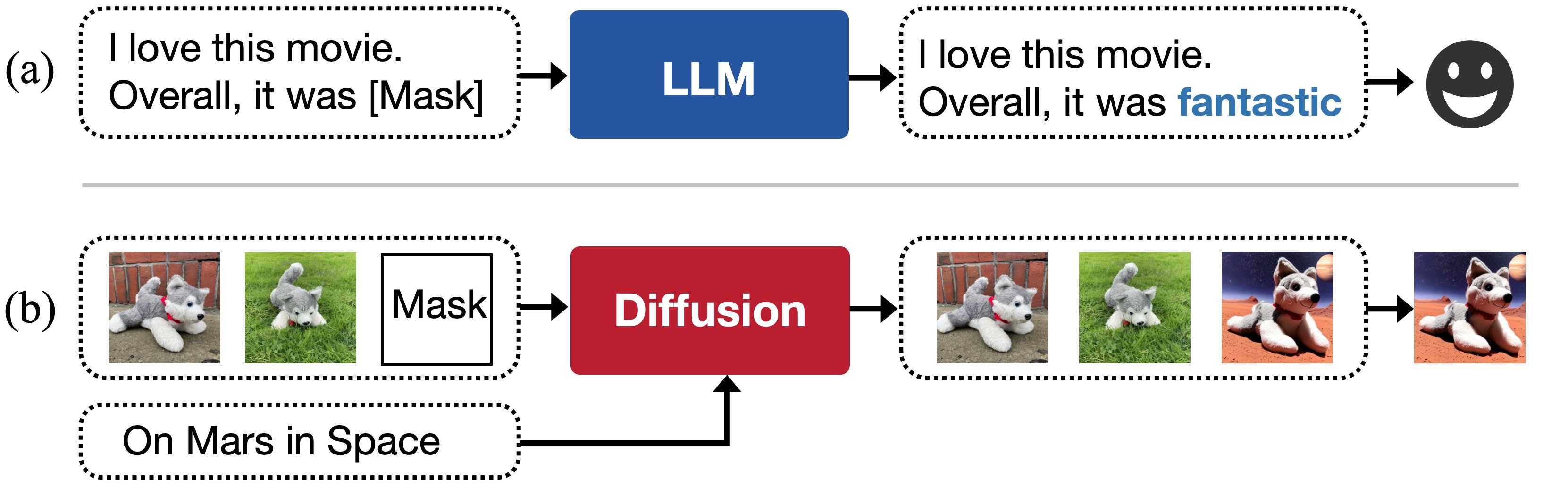}
  \caption{(a) Pretrained large language models can be applied to text classification tasks through appropriate text contexts. (b) The idea is that diffusion models can utilize visual contexts to adapt to downstream image generation tasks.}
  \label{fig3}
\end{figure}

Meanwhile, large-scale pretrained LLMs \cite{brown2020language, gao2020making} and Vision Transformers \cite{radford2021learning} have demonstrated remarkable adaptability in various downstream natural language processing and computer vision tasks. 
They retain the structure of pretrained models while transforming the downstream tasks to closely resemble the approaches used in pretraining.
As shown in Figure \ref{fig3}(a), prompt learning techniques \cite{radford2019language, raffel2020exploring, brown2020language} have emerged as an effective method to leverage tailored prompts or contexts for enabling GPT \cite{brown2020language} to excel in text classification tasks and specific text generation scenarios \cite{brown2020language, gao2020making, liu2022p}.
By structuring downstream tasks to closely resemble the pretraining process, we can effectively leverage the knowledge acquired during pretraining, 
thereby alleviating the learning difficulty of downstream tasks.

Inspired by this, we explore the potential of large pretrained diffusion models in adaptation to various downstream tasks as a DreamBooth task case shown in Figure \ref{fig3}(b). 
By constructing task-specific training data that closely resembles the pretraining setup, 
we can harness the powerful image-generative capabilities of diffusion models and further extend their applicability to various tasks.
To achieve this, we propose a novel simple, efficient, and generalizable approach named ONE-PIC. 
Motivated by the in-context tuning of LLMs above, we consider inducing the power of the visual context in pretrained diffusion models.
The visual context has shown its effectiveness in zero-shot image inpainting without training \cite{lugmayr2022repaint}.
Our previous work, StoryImager \cite{tao2025storyimager}, introduced Storyboard-based Generation, demonstrating the effectiveness of visual context to generate and complete 4-panel story images. 
In ONE-PIC, we propose In-Visual-Context Tuning to accommodate a wider range of downstream tasks.
It constructs task-specific training data by arranging source images and target images into a single image.
We transform the different fine-tuning objectives of downstream tasks to a unified masked target parts prediction.
This approach brings downstream fine-tuning closer to the pretraining process, 
allowing for better retention of pretrained knowledge and facilitating faster adaptation to various downstream tasks. 
For example, in the case of virtual try-on, our ONE-PIC achieves comparable results with only 2\% of the resources required by StableVTON \cite{kim2024stableviton}.
Additionally, our model has been adapted for various applications (see Figure \ref{fig1}), showcasing outstanding effectiveness in adapting different tasks.

Overall, our contributions can be summarized as follows:
\begin{itemize}
    \item We propose a simple, efficient, and convenient downstream fine-tuning framework that accelerates the model's adaptation to downstream tasks.
    \item We propose In-Visual-Context Tuning, a fine-tuning approach that closely resembles pretraining, enabling our model to adapt more rapidly to various downstream tasks.
    \item We introduce a Mask-based training and inference strategy that consolidates various downstream tasks into masked parts prediction.
    \item Extensive experiments on widely used datasets demonstrate that our ONE-PIC can adapt to various downstream tasks more quickly and at a lower cost.
\end{itemize}

\section{Related Work}

\subsection{Text-to-Image Synthesis}

Recent advancements in text-to-image synthesis have primarily focused on three main frameworks: Generative Adversarial Networks (GANs), Auto-regressive models, and Diffusion models.
Text-to-image GANs \cite{zhang2017stackgan, xu2018attngan, zhu2019dm, tao2020df, tao2023galip} utilize adversarial training techniques that involve a generator and a discriminator competing against each other. 
On the other hand, large-scale autoregressive models like DALL·E \cite{ramesh2021zero}, Make-A-Scene \cite{gafni2022make}, and Parti \cite{yu2022scaling} have demonstrated remarkable scalability and proficiency in image synthesis.
Diffusion models \cite{sohl2015deep, dhariwal2021diffusion, ho2020denoising, ho2022cascaded, nichol2021improved}, such as VQ-Diffusion \cite{gu2022vector}, GLIDE \cite{nichol2021glide}, DALL-E2 \cite{ramesh2022hierarchical}, Latent Diffusion Models (LDM) \cite{rombach2022high}, and Imagen \cite{saharia2022photorealistic}, have attracted significant attention in the research community. 
As likelihood-based models, they effectively mitigate the common issues of mode collapse and instability during training that are often encountered with GANs, 
enabling the generation of a more diverse range of images.

\subsection{Downstream Finetuning}

Diffusion models encompass a wide range of downstream tasks, often necessitating the design of distinct fine-tuning networks for each specific application. 
Visual conditional controls \cite{t2iadapter, li2023gligen, qin2023unicontrol, controlnet, controlnetplus} play a crucial role in these applications by providing spatial controls alongside textual conditions, allowing users to effectively manage the structure and content of generated images.
ControlNet \citep{controlnet} exemplifies this approach as it freezes the main U-Net network and incorporates a learnable encoder in parallel to extract visual condition information, 
combined with zero convolutions for more stable fine-tuning.
Subject-driven generation \citep{text_inv, ruiz2023dreambooth, li2024blip_diff} is another common need, with Dreambooth as a notable example. 
Dreambooth \citep{ruiz2023dreambooth} fine-tunes the model to bind a unique identifier to a specific subject, 
enabling the generation of new images based on just a few provided examples, thus allowing the subject to be depicted in various scenes.
Image editing \cite{Liu2020Open-Edit, Meng2022SDEdit, Brooks2022InstructPix2Pix, Nichol2022GLIDE, zhang2024magicbrush} is also a significant downstream application. 
SDEdit \citep{Meng2022SDEdit} serves as a representative work in this area as it utilizes a pretrained model to add noise to an input image and then denoise it based on a new target prompt.
Virtual try-on \citep{choi2021viton, lee2022high, morelli2023ladi, xie2023gp, kim2024stableviton} is another prevalent application in this field, requiring a more robust visual encoder to extract clothing features, 
thereby enabling accurate integration of character and clothing attributes.
Furthermore, the downstream applications of diffusion models extend to style transfer, story visualization, portrait generation, and more \cite{rahman2023make,liu2023intelligent,gong2023talecrafter}. 
These applications typically utilize their own fine-tuning frameworks, resulting in gaps in model design and posing challenges to the development of a general fine-tuning framework.

Recently, several works have attempted to propose universal fine-tuning frameworks for downstream tasks, such as Prompt Diffusion \cite{wang2023context} and OmniGen \citep{xiao2024omnigen}. 
However, Prompt Diffusion employs a framework similar to ControlNet, which limits its applicability to Visual Conditional Controls. 
OmniGen is a novel approach that introduces a new Unified Image Generation framework, utilizing a Transformer as the core generative architecture, 
directly linking text and image features to accommodate various generation tasks. 
However, OmniGen involves training a large network from scratch, resulting in significant costs. 
In contrast to these previous methods, our ONE-PIC focuses on leveraging the inherent capabilities of pretrained models. 
By constructing visual context, we enable the model to adapt quickly to downstream tasks.

\section{Method}

In this work, we propose a straightforward, efficient, and versatile approach for fine-tuning diffusion models across various downstream tasks. 
Our method, named ONE-PIC, is designed to enhance the adaptability of these models while maintaining their performance.
In the following sections, we will first provide a comprehensive overview of the ONE-PIC framework, highlighting its key components and advantages. 
We will then delve into a detailed explanation of the proposed In-Visual-Context Tuning, which leverages visual context to enhance model performance. 
Additionally, we will outline our innovative Masking Strategy, which consolidates multiple tasks into a unified prediction framework. 
Our ONE-PIC not only simplifies the fine-tuning process but also ensures that the models retain their pretraining knowledge while effectively adapting to new tasks.

\begin{figure*}[t] \small
    \centering
    \includegraphics[width=1\linewidth]{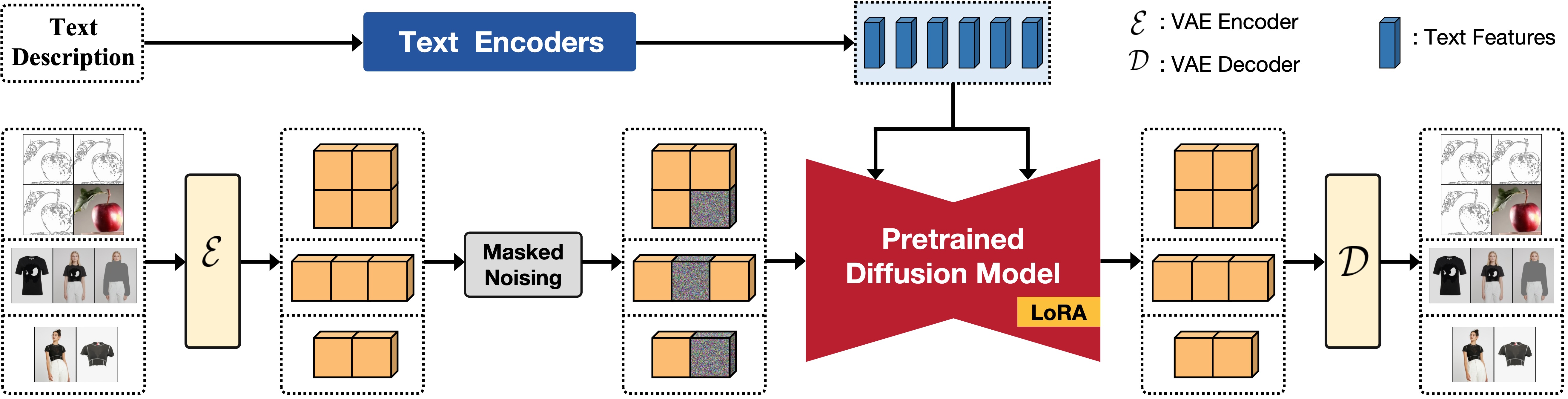}
    \caption{The architecture of ONE-PIC for different downstream tasks. We show the visual context of visual conditional control, virtual try-on, and fashion extraction. }
    \label{fig4}
  \end{figure*}

\subsection{Model Overview}

As shown in Figure \ref{fig4}, our proposed ONE-PIC has a simple framework.
It comprises a pretrained Text Encoder, a pair of Image Encoder $\mathcal{E}$ and Decoder $\mathcal{D}$ from a pretrained autoencoder, 
a pretrained diffusion model \cite{rombach2022high} with Low-Rank Adaptation (LoRA) \cite{peft}, and a Masking Strategy.
Our ONE-PIC fundamentally inherits the framework of pretrained SDXL \cite{podell2023sdxl} without adding any extra modules. 
We have only introduced an additional masking strategy to perform a masked noising process on the input visual context and target image, 
allowing for adaptation to various downstream fine-tuning tasks.

The input images are initially encoded into latent space, while the text encoder converts the text descriptions into text features. 
Next, ONE-PIC constructs the visual context from the latent features of the input reference images and merges this visual context with the latent features of the target image to form a comprehensive set of latent features.
During training and inference, the Masking Strategy is employed to mask the latent features of the target images. 
The U-Net of the pretrained diffusion model then processes these latent features alongside the text features and fuses them together. 
The entire model is trained to predict the noise present in the masked regions. 
Finally, after reversing the diffusion process over multiple steps, the latent features are decoded into the target images.

\subsection{In-Visual-Context Tuning}

Since pretrained diffusion models typically use only text as a condition, 
existing downstream fine-tuning methods often consider adding branch networks to adapt these models for various generation tasks. 
Among these, ControlNet \citep{controlnet} and IP-Adapter \cite{ye2023ipadapter} are particularly representative. 
ControlNet directly copies the encoder of the diffusion model's U-Net to encode the input image conditions, 
while IP-Adapter introduces an additional CLIP Image Encoder to capture image features.
ControlNet focuses more on the spatial information of the images, whereas IP-Adapter, by utilizing highly encoded visual features, emphasizes high-level information. 
An effective universal fine-tuning framework should be capable of capturing both the intricate details of images and their high-dimensional semantics to accommodate different downstream tasks.

Recently, large-scale pretrained models have shown impressive zero-shot capabilities across a variety of tasks in both natural language processing and computer vision. 
These models retain the structure of pretrained architectures while adapting downstream tasks to closely align with the pretraining methods.
For instance, prompt learning \cite{radford2019language, raffel2020exploring, brown2020language} utilizes the creation of appropriate prompts or contexts, 
enabling GPT \cite{brown2020language} to perform tasks such as text classification and specific text generation \cite{brown2020language, gao2020making, liu2022p}. 
The CLIP model \cite{radford2021learning} achieves zero-shot image classification by constructing appropriate text prompts.
By structuring downstream tasks to closely mirror the pretraining process, we can more effectively leverage the knowledge acquired during pretraining.

Inspired by this, we explore the potential of large pretrained diffusion models. 
We find that the images generated by diffusion models possess rich details and accurate high-level visual semantics. 
Thus, a question arises: Can we leverage the strong visual feature extraction capabilities of diffusion models to adapt to various downstream generation tasks?
Motivated by it, we propose In-Visual-Context Tuning. 
This approach transforms reference images into a visual context, which is then combined with the target image to form a complete representation. 
Unlike methods that add additional networks, this complete image can be directly applied in the pretrained framework without the need for additional network design.
It forms the foundation for constructing a unified framework for various downstream applications.
This unified framework allows us to seamlessly integrate the strengths of diffusion models while minimizing the complexity often associated with task-specific adaptations. 
By utilizing In-Visual-Context Tuning, we can capture both the fine-grained details and the high-level semantics of the images, enabling the model to effectively understand and generate content based on diverse inputs.

\begin{figure}[t] \small
  \centering
  \includegraphics[width=0.99\linewidth]{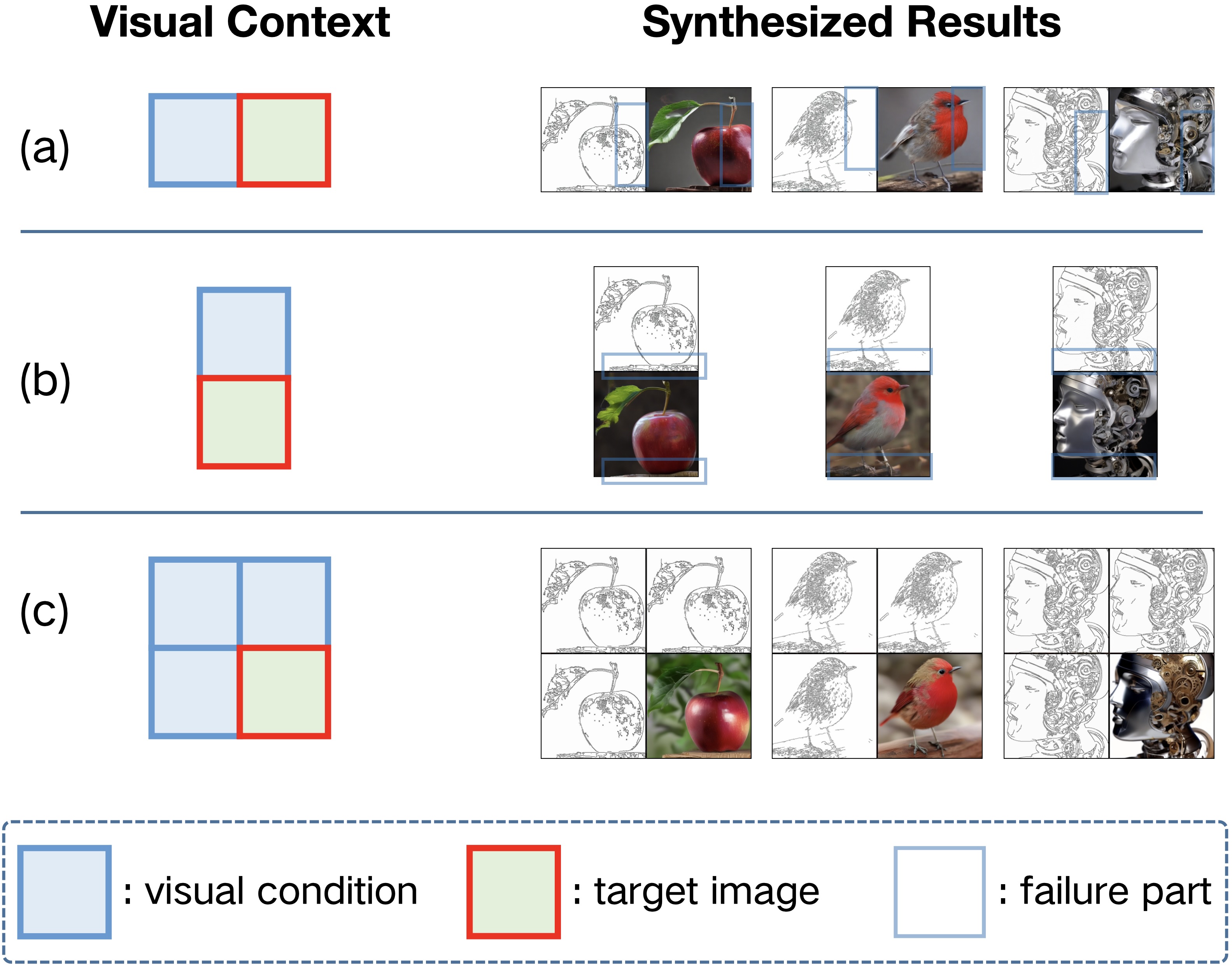}
  \caption{In visual conditional control generation tasks, different visual contexts can have varying impacts on the results.}
  \label{fig31}
\end{figure}

Similar to prompt learning \cite{radford2019language, raffel2020exploring, brown2020language} in language models, we found that combining images into different visual contexts produces varying effects. 
Taking the ControlNet task as an example, we compared several different visual contexts, with the results shown in Figure \ref{fig31}. 
We observed that using inputs shaped as \(1 \times 2\) or \(2 \times 1\) resulted in slight distortions and stretching along both the horizontal and vertical axes. 
However, when using inputs shaped as \(2 \times 2\), these issues did not occur.
This may be related to the task's requirement for precise positional alignment. 
Inputs shaped as \(2 \times 2\) can provide more positional information along both the horizontal and vertical axes of the target image, thereby adapting better to this task. 
We also observed that the likelihood of errors occurring in the central areas of the image is significantly lower than that in the edges. 
This may be attributed to a pretraining bias, as the content in the center of most images is generally more important than that at the edges.
These observations led us to adopt different visual contexts when adapting to various downstream tasks, as illustrated in Figure \ref{fig4}.
We summarize additional design insights in the experimental section.

\begin{figure*}[t] \small
    \centering
    \includegraphics[width=1\linewidth]{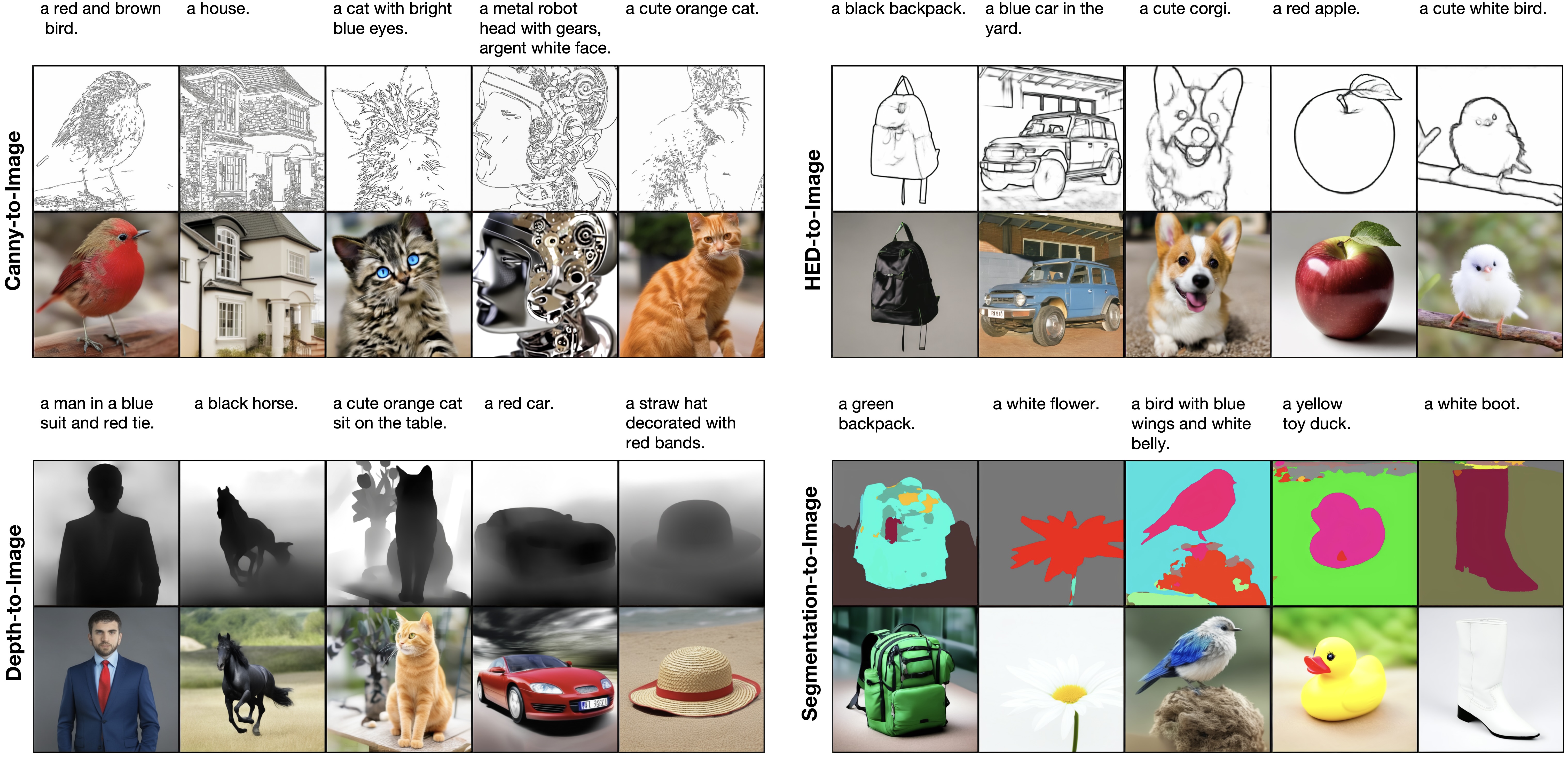}
    \caption{The synthesized results of ONE-PIC under different visual conditional controls.}
    \label{fig5}
  \end{figure*}

\subsection{Masking Strategy}

To enable the model to reference the visual context when generating the target image, we propose a Masking Strategy for the training and inference stage. 
Unlike directly adding noise to the entire image, our ONE-PIC applies noise solely to the target area, preserving the clean visual context for effective image feature extraction. 
This approach closely resembles image inpainting. 
Interestingly, we find that image inpainting capabilities appear to be inherently embedded within pretrained diffusion models, 
as demonstrated by the image inpainting model \cite{lugmayr2022repaint}, which can perform inpainting without requiring fine-tuning of the diffusion model. 
By adopting this strategy, we can better leverage the knowledge embedded in pretrained diffusion models, thereby accelerating adaptation to downstream tasks.

In the forward process, the Masking Strategy samples a binary mask \( m \) to indicate the target region for generation. 
It then applies a masked noising process to the latent features \( x \). We define \( x_0 = x \) and only add noise to the latent of the target images, rather than the entire latent:
\begin{align}
    &\tilde{x}_t = \sqrt{\bar{\alpha}_t} x_0 + \sqrt{1 - \bar{\alpha}_t} \epsilon,\\
    &x_t = \tilde{x}_t \odot m + x_0 \odot (1 - m),
\end{align}
where \( \epsilon \sim \mathcal{N}(\mathbf{0}, \mathbf{I}) \) and \( t \) denotes the timestep in the forward process. 
The Masking Strategy enables the model to utilize visual information from the given reference images and learn to recover the masked parts \( x_0 \odot m \). 
This ensures that the generated target images within the mask \( m \) are consistent with the provided reference images. 
Following the approach in \cite{ho2020denoising}, we train a network \( \epsilon_{\theta} \) to predict the noise \( \epsilon \) from the noisy \( x_t \):
\begin{align}
    \mathcal{L}_{\text{DM}} = \mathbb{E}_{ \epsilon \sim \mathcal{N}(0, I)}\left[\|m \odot\epsilon - m \odot \epsilon_\theta(x_t, t, e)\|_2^2\right].
\end{align}
where \( e \) represents the text description. 
The typical training scheme used in stable diffusion predicts the loss over the entire image latent. 
However, since our goal is to predict the masked target parts based on the provided reference images and text descriptions, 
we only compute the loss for the target masked portions.

During the inference phase, we apply a masked noising process on the target latent features \( x_T = \epsilon \odot m + x_0 \odot (1 - m) \), where \( T \) is the number of sampling steps. 
The unmasked parts remain intact throughout the denoising process at each step. 
Subsequently, we reverse the diffusion process to obtain the completed latent feature \( x_0 \).

The significance of the Masking Strategy for training and inference lies in its ability to enhance the model's focus on relevant visual information while generating images. 
By selectively applying noise only to the target area, the model can maintain the integrity of the visual context, which is crucial for effective image feature extraction.

\section{Experiments}

\begin{figure}[t] \small
  \centering
  \includegraphics[width=\linewidth]{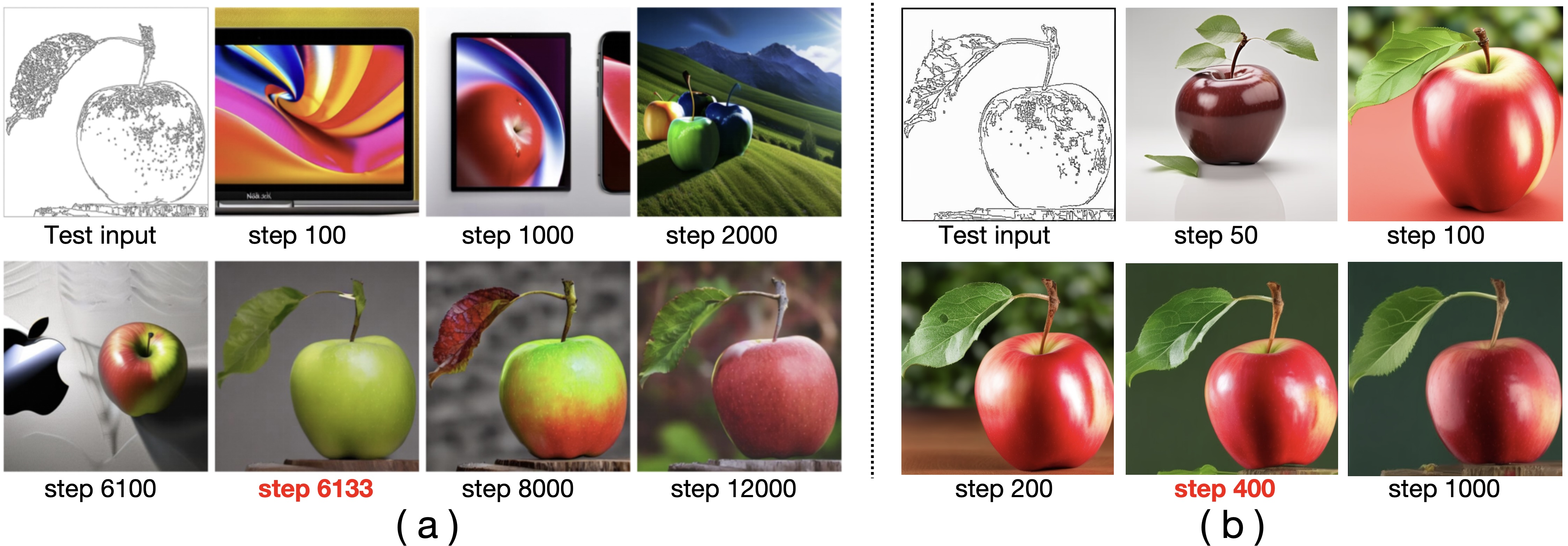}
  \caption{(a) The convergence process of ControlNet\cite{controlnet}. (b) The convergence process of ONE-PIC. The convergence of our ONE-PIC is significantly faster than that of ControlNet.}
  \label{fig51}
\end{figure}

\begin{figure*}[t] \small
    \centering
    \includegraphics[width=\linewidth]{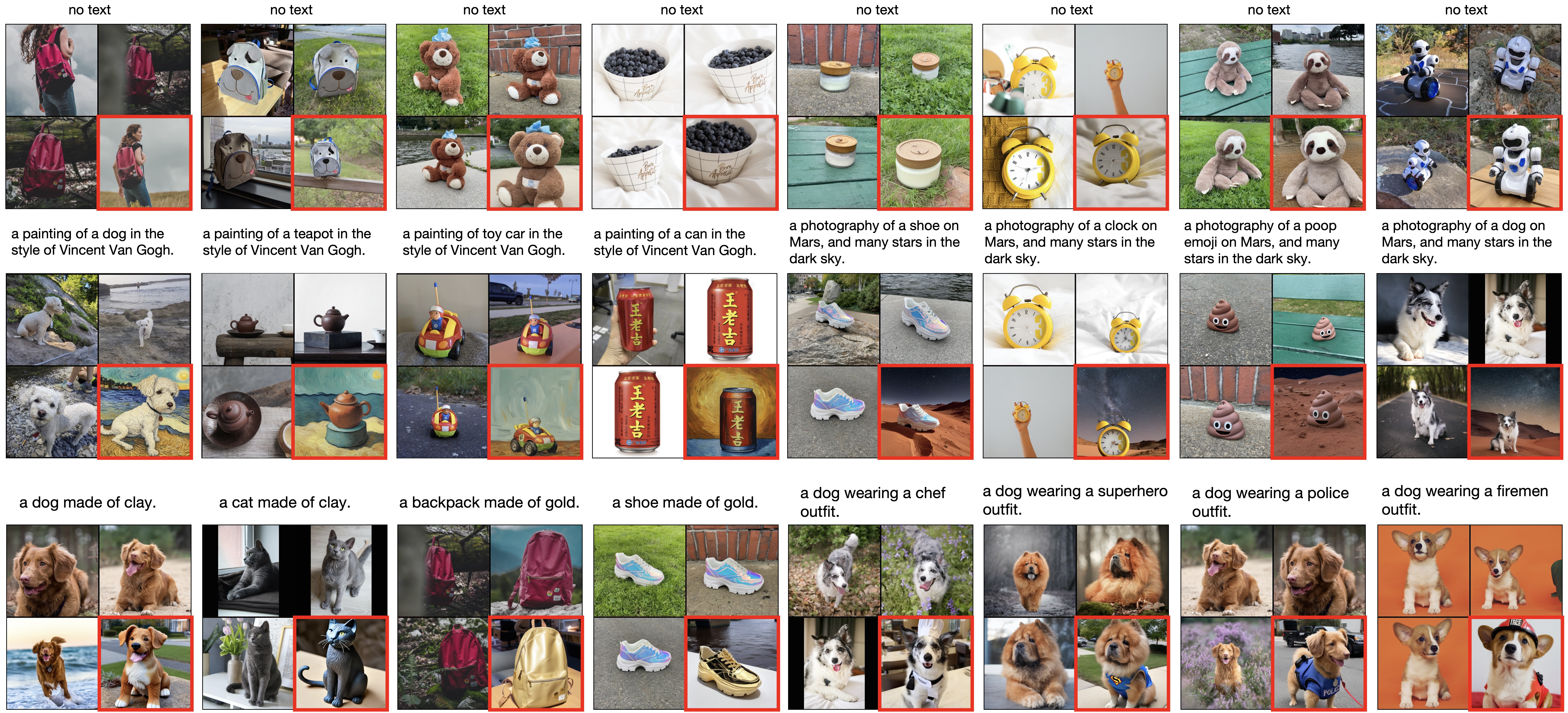}
    \caption{The synthesized images (in red box) of ONE-PIC under DreamBooth \citep{ruiz2023dreambooth} task which generates new images according to 3 reference images on top-left and given text prompts.}
    \label{fig6}
  \end{figure*}

To validate the quality of images generated by ONE-PIC and its learning efficiency, 
we conducted extensive experiments across multiple tasks and performed detailed analysis on four common downstream generation tasks: visual conditional controls, Dreambooth, image editing, and virtual try-on. 
For each downstream task, we first introduce the dataset and training details, 
followed by an analysis of experimental results.
Finally, we present some generation results of ONE-PIC across additional tasks to demonstrate its adaptability to a variety of different applications.

Our method is based on the SDXL \cite{podell2023sdxl} model.
We freeze the pretrained autoencoder and text encoder and finetune the U-Net through LoRA with $\alpha=4$, $r=32$. 
We select the ``attn1.to\_q'', ``attn1.to\_k'', and ``attn1.to\_v'' layers in self-attention for LoRA fine-tuning.
Due to the high efficiency of learning in ONE-PIC, 
the learnable parameters required for downstream fine-tuning account for only 0.618\% of the model, 
yet it achieves performance comparable to current fine-tuning models.
We use the AdamW optimizer to train our model.
We set the learning rate $0.001$. 
All models were trained on 4 $\times$ NVIDIA RTX A6000 GPUs.

\subsection{Visual Conditional Controls}

Visual Conditional Controls play a crucial role in the downstream applications of diffusion models by providing spatial controls beyond textual conditions, 
enabling users to better manage the structure and content of generated images. 
For this task, we trained our ONE-PIC on a randomly selected subset of 20,000 samples from the LIAON-Art dataset. 
The images generated by our ONE-PIC are displayed in Figure~\ref{fig5}.
Our ONE-PIC effectively captures the spatial information contained within visual conditional controls, 
generating realistic images that correspond to this spatial data while ensuring semantic consistency with the textual conditions. 
Furthermore, we compared the learning efficiency of our model with that of the widely used ControlNet \cite{controlnet}. 
The convergence processes for both ControlNet and ONE-PIC are illustrated in Figure~\ref{fig51}. 
We utilized the convergence process images from the ControlNet paper for this comparison.
Our experiments demonstrated that after just 400 training steps, which took only 10 minutes, our model had already grasped the fine-tuning task. 
In contrast, ControlNet requires 6,000 steps to understand the task and necessitates several days of training for complete convergence. 
Thus, ONE-PIC demands significantly less data and time, showcasing its efficiency and effectiveness in adapting to downstream tasks.
This makes our fine-tuning strategy more accessible to a broader range of users.

\begin{figure*}[t] \small
    \centering
    \includegraphics[width=1\linewidth]{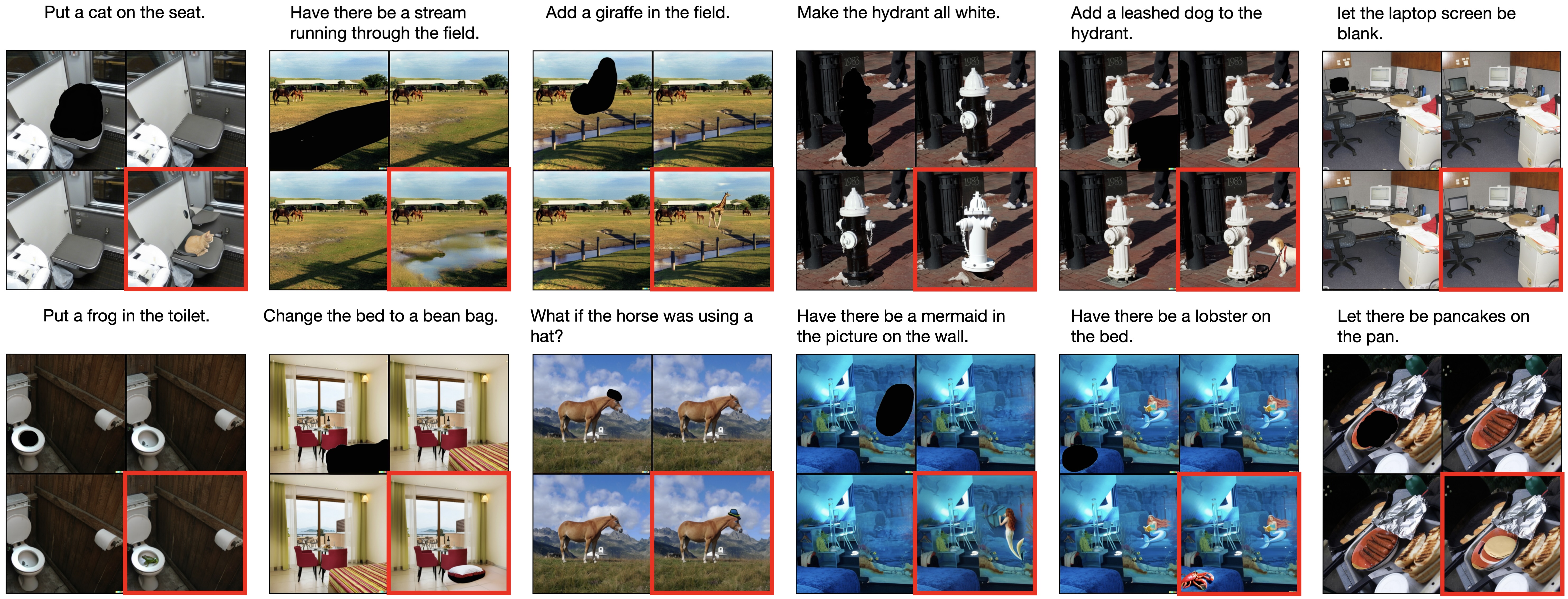}
    \caption{The synthesized results (in red box) of ONE-PIC under different editing requirements. The top-left corner contains the masked image, indicating the area that needs to be edited, while the top-right and bottom-left corners display the source images.}
    \label{fig7}
  \end{figure*}

\subsection{DreamBooth}

The goal of Dreambooth \citep{ruiz2023dreambooth} is to generate new images of a subject based on just a few provided images, allowing the subject to be depicted in various scenes. 
Following previous work \citep{xiao2024omnigen, ruiz2023dreambooth}, we evaluate the subject-driven generation capability using DreamBench, 
which consists of 750 prompts for 30 different subjects (e.g., dogs, cats, and toys).
We provided the model with three reference images, positioned in the top-left corner of a larger image
while placing the target image in the bottom-right corner. We trained for 10,000 steps on the DreamBench dataset, which took approximately 2.5 hours. 
The images generated by our ONE-PIC are shown in Figure~\ref{fig6}.
To assess ONE-PIC's generalization capabilities, we prompted it to generate a variety of creative images, such as ``on Mars'', ``made of clay'', and ``made of gold''. 
The results demonstrate that our model effectively captures the features of the subjects in the reference images and generates new, 
realistic images that align with the textual descriptions in diverse scenes.

\begin{figure*}[t] \small
    \centering
    \includegraphics[width=1\linewidth]{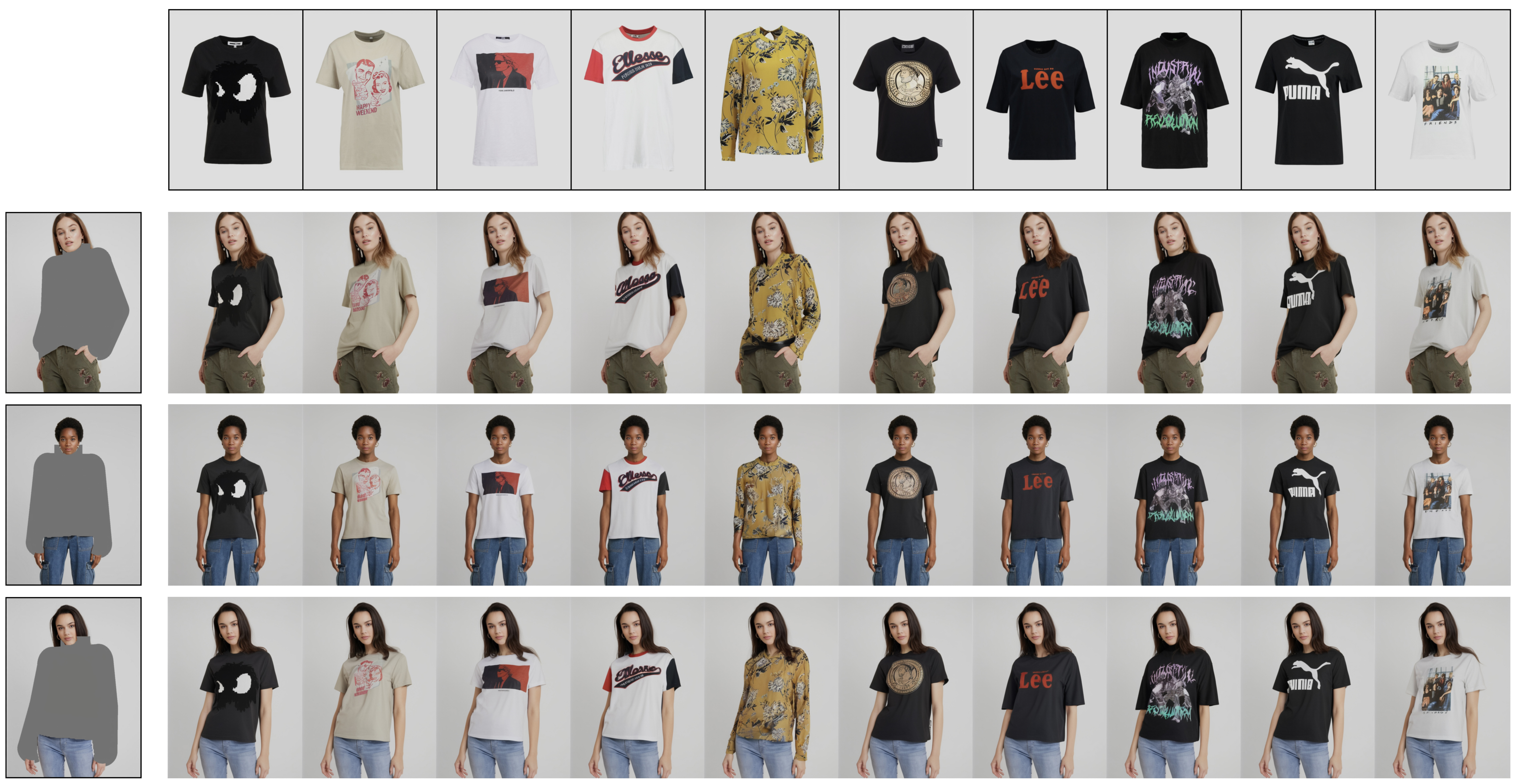}
    \caption{The synthesized results of ONE-PIC under virtual try-on tasks.}
    \label{fig8}
  \end{figure*}

\begin{figure*}[t] \small
  \centering
  \includegraphics[width=\linewidth]{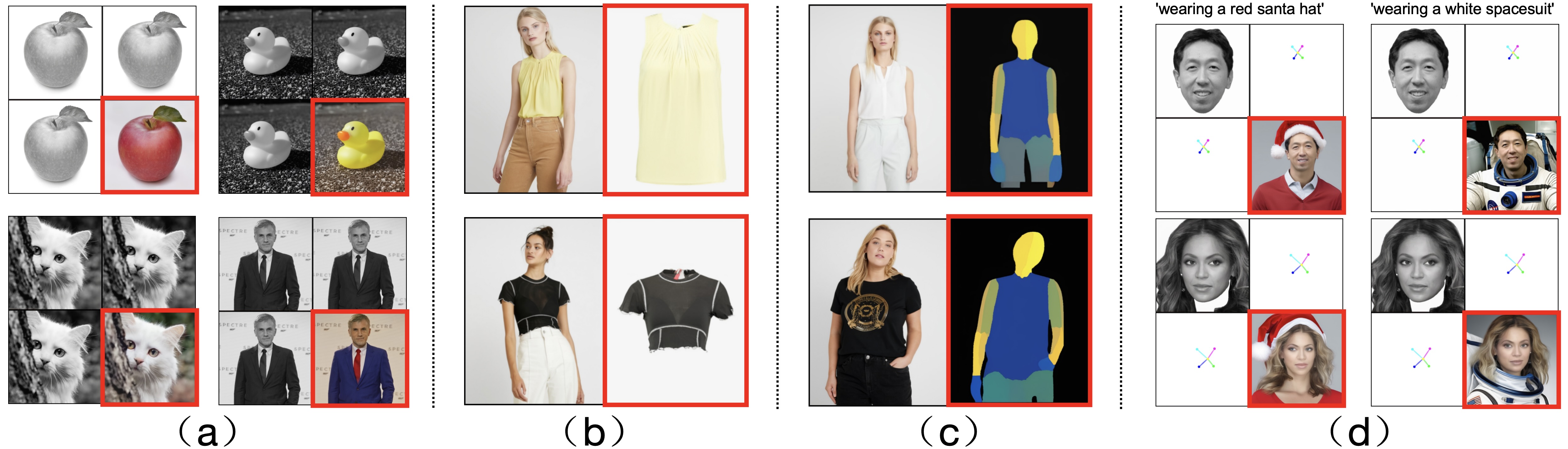}
  \caption{The visual context and generated results of Image Colorization(a), Fashion Extraction(b), Image Segmentation(c), and Identity-Preserved Portrait Generation(d).}
  \label{fig9}
\end{figure*}

\subsection{Image Editing}

We utilized the MagicBrush dataset \cite{zhang2024magicbrush} to train and evaluate our model for image editing. 
Reference images were placed in three positions at the top-left corner of a larger image, while the target image was positioned in the bottom-right corner. 
For the mask-given image editing scenario, we replaced the images in the top-left corner with those containing the mask information.
We trained for 10,000 steps on the MagicBrush training dataset, which took approximately 2.5 hours. 
The edited images generated by our ONE-PIC are shown in Figure~\ref{fig7}. 
To assess ONE-PIC's editing capabilities, we tested the model on a variety of editing tasks, 
such as ``put a cat on the seat'', ``make the hydrant white'', and ``change the bed to a bean bag''.
The results demonstrate that our ONE-PIC can accurately edit images according to the provided textual instructions.

\subsection{Virtual Try-On}

Following previous work \cite{kim2024stableviton}, we adopt the VITON-HD \cite{choi2021viton} dataset to train and evaluate our model for Virtual Try-On.
We found that for the Virtual Try-On task, the $1\times3$ layout performed better than the $2\times2$ layout. 
In the $1\times3$ configuration, we placed the clothing image in the left panel, the masked person image in the right panel, and the target image in the center. 
This arrangement yielded the best results.
We resized the images to a resolution of $512\times384$.
We trained for 10000 steps on the VITON-HD training dataset, which took approximately 2 hours.
We show the generated images of virtual try-on synthesized by our ONE-PIC in Figure~\ref{fig8}.
From the generated results, our ONE-PIC accurately produces fitting images of clothing that align with the model's pose. 
It ensures the realism of the generated images while preserving details of the clothing, such as the prints and textures on the clothing.

\subsection{More Downstream Tasks}

In addition to the four common downstream tasks, we successfully adapted ONE-PIC to several more tasks, 
including Image Colorization, Fashion Extraction, Image Segmentation, and Identity-Preserved Portrait Generation, etc.
We show some results in Figure~\ref{fig9}.
Each task only required about two hours for fine-tuning. 
Notably, Fashion Extraction is a newly proposed task that involves extracting clothing from a model's photo and organizing it into another image. 
Based on our proposed ONE-PIC, no intricate network design and professional knowledge is needed, simply by stitching images together and applying a mask to the clothing image, 
ONE-PIC can be extended to the Fashion Extraction task.
In the future, we will adapt ONE-PIC to more downstream tasks.

\subsection{Design Tricks for Visual Context}
In our proposed In-Visual Context Tuning, we discovered that the visual contexts needed for different tasks can vary significantly. 
This phenomenon is akin to how different text prompts can lead to distinct outcomes in generative language models. 
Drawing on the extensive insights gained from our experiments, we have compiled a set of design tricks for crafting effective visual contexts. 
These design tricks are summarized as follows:
\begin{itemize}[leftmargin=*]
    \item If your downstream generation task requires precise positional control of the generated images, similar to that of visual conditions in tasks like ControlNet, 
    it is advisable to use $2 \times 2$ shaped inputs. This configuration allows the condition images placed to the left and right of the target image to provide horizontal positional information, 
    while the condition images positioned above and below the target image supply vertical positional information. 
    This arrangement aids the network in better learning spatial positional cues.
    \item If your downstream generation task involves multiple different conditions, it is beneficial to place the target images in the center position. 
    This is due to the prior belief that the visual content in the central region may be more important for pretrained models. 
    Positioning the target image in the center can facilitate faster convergence for the model.
    \item If your downstream generation task provides rich visual information, as seen in the source images of image editing tasks, 
    and requires faster inference speed, you might consider using $1 \times 2$ shaped inputs or $2 \times 1$ shaped inputs. 
    While switching to more robust $2 \times 2$ shaped inputs could potentially yield better results, the improvement may not be very significant. 
    In fact, using $1 \times 2$ or $2 \times 1$ shaped inputs can facilitate quicker inference, as the size of the input features is reduced.
\end{itemize}

\subsection{Limitations and Discussions}
Our ONE-PIC shows superiority in downstream fine-tuning, but some limitations should be considered in future studies. 
Our ONE-PIC introduces visual context, which results in a greater computational burden during the inference process. 
Second, the diffusion models based on the DiT \cite{peebles2023scalable} architecture have demonstrated strong capabilities. 
In the future, we plan to introduce strategies that can reduce computational costs, and adapt a pretrained diffusion model based on DiT \cite{peebles2023scalable}.

\section{Conclusion}

We present a simple, efficient, and convenient downstream fine-tuning framework, named ONE-PIC, that accelerates the model's adaptation to various tasks. 
Our In-Visual-Context Tuning approach closely mirrors pretraining, allowing for faster adaptation to different downstream applications. 
Additionally, we introduce a Masking Strategy for training and inference that consolidates multiple downstream tasks into predictions of masked portions. 
Extensive experiments demonstrate that our ONE-PIC can adapt more quickly and at a lower cost to a range of downstream tasks.


{
    \small
    \bibliographystyle{ieeenat_fullname}
    \bibliography{main}
}



\end{document}